\documentclass{article}

\usepackage[nonatbib,final]{nips_2017}

\usepackage{tabulary}
\usepackage{graphicx}
\usepackage[font={small,it}]{caption}
\usepackage{relsize}
\usepackage{amsmath}
\usepackage{sidecap}
\usepackage{url}
\usepackage{float}
\usepackage{array}
\usepackage{multirow}
\usepackage{pdfpages}
\usepackage{enumitem}

\usepackage{rotating}
\usepackage{color, colortbl}
\usepackage{hhline}
\usepackage{adjustbox}
\usepackage{hyperref}

\newcolumntype{R}[2]{%
    >{\adjustbox{angle=#1,lap=\width-(#2)}\bgroup}%
    l%
    <{\egroup}%
}

\definecolor{Gray}{gray}{0.9}

\title{False Positive and Cross-relation Signals in Distant Supervision Data}
\author{
  Anca Dumitrache \\
  Vrije Universiteit Amsterdam, \\
  CAS IBM Netherlands \\
  \texttt{anca.dumitrache@vu.nl} \\
  \And
  Lora Aroyo \\
  Vrije Universiteit Amsterdam \\
  \texttt{lora.aroyo@vu.nl} \\
  \And
  Chris Welty\\
  Google Research, New York \\
  \texttt{cawelty@gmail.com} \\
}

\begin{document}

\maketitle

\begin{abstract}
Distant supervision (DS) is a well-established method for relation extraction from text, based on the assumption that when a knowledge-base contains a relation between a term pair, then sentences that contain that pair are likely to express the relation. In this paper, we use the results of a crowdsourcing relation extraction task to identify two problems with DS data quality: the widely varying degree of false positives across different relations, and the observed causal connection between relations that are not considered by the DS method. The crowdsourcing data aggregation is performed using ambiguity-aware CrowdTruth metrics, that are used to capture and interpret inter-annotator disagreement. We also present preliminary results of using the crowd to enhance DS training data for a relation classification model, without requiring the crowd to annotate the entire set.
\end{abstract}

\section{Introduction}

Distant supervision (DS)~\cite{mintz2009distant,Welty:2010:LSR} is a well-established semi-supervised method for performing relation extraction from text. It is based on the assumption that, when a knowledge-base contains a relation between a pair of terms, then any sentence that contains that pair is likely to express the relation. This approach can generate false positives, as not every mention of a term pair in a sentence means a relation is also expressed~\cite{DBLP:conf/ijcai/FengGQLL17}. Furthermore, dependencies between the semantics of the relations such as causality or contradiction are also not considered by the DS methodology. It is often assumed that these disadvantages are compensated for by the scale of the data a DS method can produce, or can be largely overcome with crowdsourced human annotation~\cite{angeli2014combining}.  

In this paper we identify two specific problems we have found with distant supervision training data: the widely varying degree of false positives across different TAC-KBP relation types, and the observed causal connection between relations. We expose these problems using a novel approach to gathering human annotated data, CrowdTruth~\cite{aroyo2014threesides,aroyo2015truth,aroyo2013crowd}, analyze them, and offer preliminary heuristic and statistical approaches to incorporating them back into DS-based training, that provides better sentence-level relation extraction results, without requiring crowdsourcing on the full set of data.

\section{Background}
In recent years, researchers have explored unsupervised methods for correcting DS data. For the task of knowledge base completion, \cite{DBLP:conf/ijcai/FengGQLL17} applied memory networks both to correct false positives in the data, and to capture dependencies between relations. For the same task, \cite{Jiang2016RelationEW} developed a loss function that works with multi-label data, in order to capture co-occurring relations. For relation classification from sentences, \cite{DBLP:conf/acl/SantosXZ15} learn embeddings that capture cross-signals between relations. However, these approaches are dependent on training data that can express relation semantics with at least some accuracy. The initial experiments presented in this paper show the error rate in the DS data can be so high that unsupervised learning becomes unreliable when it comes to capturing cross-relation signals.

Crowdsourcing is a well-used approach to correcting the mistakes in DS by scaling out cheap human annotation.  We have been studying the problem of collecting human annotations from the crowd using the CrowdTruth methodology \cite{aroyo2013crowd}.  Our method differs in that it gathers many annotations for the same examples, to better reflect properties like ambiguity, human error and spam, and the target semantics \cite{aroyo2014threesides}. We have used it successfully to improve DS results for the task of medical relation extraction~\cite{DBLP:journals/corr/DumitracheAW17}, achieving annotation quality equivalent to that provided by medical experts, at less than half the cost.

\section{Data and Annotation with CrowdTruth}

We began by having the crowd annotate 2,500 sentences from the NIST TAC-KBP 2013 English Slotfilling data that were annotated with DS.  We split the data in half into a test and dev set. We focused the crowd annotations on a subset of 16 relations (Fig.\ref{fig:fp_rate}) that occur between terms of types $Person$, $Organization$ and $Location$. We ran a multiple-choice crowdsourcing task on Crowdflower,\footnote{\url{http://crowdflower.com}} asking {\em 15 workers to annotate each sentence} with the appropriate relations, or choose the option $NONE$ if none of the relations presented apply. Workers are encouraged to select all relations that apply. Each worker was paid \$0.05 per sentence. The data is available online.\footnote{\url{https://github.com/CrowdTruth/Open-Domain-Relation-Extraction}} 

Traditionally, crowdsourcing annotations are aggregated by looking at the consensus of the workers, through methods such as majority vote. This process is based on a simplified notion of truth, under the assumption that a single right annotation exists for each example. However, in reality, truth is not universal and is strongly influenced by a variety of factors, such as the ambiguity that is inherent in natural language, resulting in artificial data that does not represent the diverse perspectives of the crowd.

To address this issue, we proposed the CrowdTruth metrics\footnote{\url{https://git.io/v5iTB}} of aggregating crowdsourcing data while also capturing and interpreting inter-annotator disagreement. The main basis for measuring quality in CrowdTruth is the triangle of disagreement, which links together sentences, workers, and relations. It allows us to assess the quality of each worker, the clarity of each sentence, and the ambiguity, similarity and frequency of each relation. The triangle model expresses how ambiguity in any of the components of the triangle disseminates and influences the other components. For example, an unclear sentence or an ambiguous relation would cause more disagreement between workers, and thus, both need to be accounted for when measuring the quality of the workers, which is then used to identify the spam workers. Among the CrowdTruth measures discussed in this paper, we calculate the per-relation false positive (FP) rate and the {\it causal power} between relation pairs (RCP) over the dev set. Spam removal was performed as well, but the details of this process are not relevant for the paper.


For every sentence, the annotations of each worker form a binary {\it worker vector}, where the relations selected are equal to `1', and the rest to `0'. The {\it sentence vector} is the sum of all worker vectors for the given sentence. Then for each relation, we compute the {\it sentence-relation score} of relation $i$ ($SRS_i$) as the ratio of workers that picked that relation over the total of number of workers. The SRS measures how clearly the relation is expressed in the sentence, and is used as a continuous truth measure in a lot of our work.  In order to make our results compatible with the AKBC community's discrete truth metrics (e.g. P, R, F1), we have chosen a threshold of 0.5 per relation, corresponding to the majority vote, that allows for multiple relations to be considered correct in a sentence.  False positive rates are then computed per relation on the dev set with this threshold.

Causal power~\cite{cheng1997causalpower} is an estimate of the probability that the presence of one relation implies the presence of another. Given two relations $i$ and $j$, $ RCP(R_i, R_j) = [ P(R_{j} | R_{i} ) - P(R_{j} | \neg R_{i} ) ] / [1 - P(R_{j} | \neg R_{i} )]$, where $P(R_{i})$ is the probability that relation $R_i$ is annotated in the sentence. This probability can be calculated on a micro basis giving us the probability of one worker annotating two relations together; the {\it macro RCP} calculates the probabilities in the sentence vectors, capturing causality as a result of two relations being annotated together in the same sentence, but not necessarily by the same workers.  We found micro RCP to be vastly inferior to macro RCP, which is further evidence of the value of having multiple workers per sentence, and only include the macro RCP results in this paper.  

In addition to the dev and test sets, we also used a training set of 235,000 sentences annotated by distant supervision from freebase relations, used in ~\cite{riedel2013universalschema}.  This data was used as a baseline for training a relation extraction classifier based on ~\cite{nguyen2015relation}.  The model is a convolutional neural network, adapted to be both multi-class and multi-label -- we use a sigmoid cross-entropy loss function instead of softmax cross-entropy, and the final layer is normalized with the sigmoid function instead of softmax -- in order to make it possible for more than one relation to hold between two terms in one sentence.
To evaluate the relation extraction model on CrowdTruth data with discrete AKBC metrics, we set a comparable threshold of 0.5 on the model confidence score, separating between negative and positive labels. However, the loss function is not computed using binary labels generated by the threshold, but using the continuous labels in both the train and test sets.

\section{Analysis}

\begin{figure}[htb!]
\centering
\includegraphics[width=0.75\textwidth]{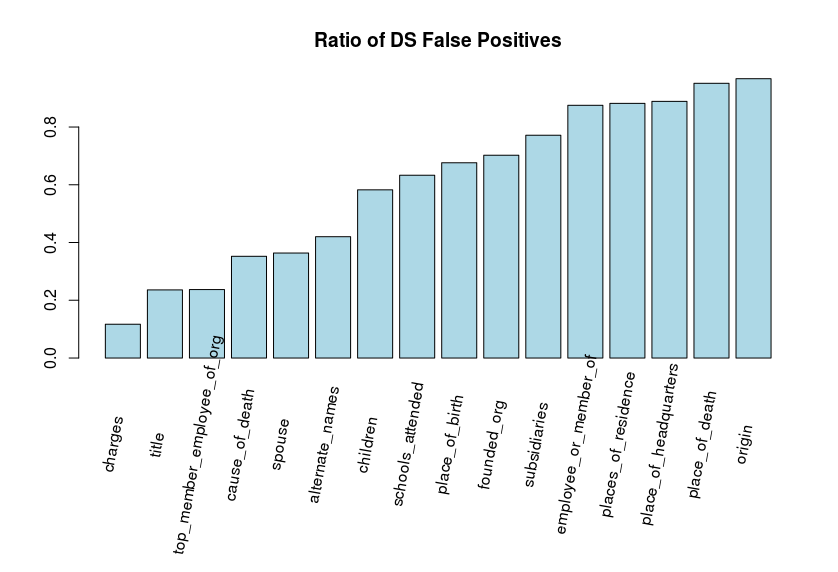}
\caption{DS ratio of false positive over all positive labels, using the crowd as ground truth.}
\label{fig:fp_rate}
\end{figure}

Using the SRS as a ground truth at a 0.5 threshold, Fig.~\ref{fig:fp_rate} shows the correctness of the DS labels on the dev set.  There is {\it considerable variation in DS data quality across relations}. The $origin$ and $place\_of\_death$ relations scored particularly badly, with more than 90\% false positives. With such a high error rate in some relations, it is arguable that any classifier could learn anything meaningful, regardless of algorithm or quantity of data.

Manual error analysis on the dev set showed that many sentences contain a $Person$ - $Location$ pair, where freebase specified both that the person resided in and died at that location. This makes intuitive sense, people tend to die in the places they live.  In most of these cases, the sentence expressed only the $places\_of\_residence$ relation, leading to the false positives. The $origin$ relation data suffers from the same problem.  Table~\ref{tab:ds_fp_ex} in the Appendix shows several examples of these sentences. This led us to consider a heuristic solution to this problem as a headroom study as well as a statistical solution. Both are discussed in Section 5.

\begin{table}[htb!]
\caption{RCP for relation subset: $place\_of\_birth$ (PoB), $origin$ (O), $places\_of\_residence$ (PoR), $place\_of\_death$ (PoD), $founded\_organization$ (FO), $employee\_or\_member$ (EoM), $top\_employee\_or\_member$ (TEoM). The scores show the causal power $RCP(R_i, R_j)$ of relations $R_i$ in the rows, over the relations $R_j$ in the columns. Significant changes between crowd annotation based causal power and distant supervision are in bold.}
\label{tab:rcp}
\scalebox{0.7}{
\begin{minipage}{.4\textwidth}
	\caption{Crowd-based RCP}
    \label{tab:crowd_rcp}
    \begin{tabular}{|r|ccccccc|}
    \hline
          & PoB   & O     & PoR   & PoD   & FO  & EoM     & TEoM \\ \hline
     PoB  & 1     & {\bf 0.64}  & 0.17  & -0.12 & -0.19 & -0.2  & -0.21  \\ 
     \cellcolor{Gray}O    & \cellcolor{Gray}{\bf 0.88}  & \cellcolor{Gray}1     & \cellcolor{Gray}0.31  & \cellcolor{Gray}-0.16 & \cellcolor{Gray}-0.29 & \cellcolor{Gray}-0.22 & \cellcolor{Gray}-0.22 \\ 
     PoR  & 0.42  & {\bf 0.56}  & 1     & {\bf -0.1}  & -0.59 & 0.12  & 0.13 \\ 
     \cellcolor{Gray}PoD  & \cellcolor{Gray}-0.03 & \cellcolor{Gray}-0.03 & \cellcolor{Gray}-0.01 & \cellcolor{Gray}1     & \cellcolor{Gray}-0.04 & \cellcolor{Gray}-0.05 & \cellcolor{Gray}-0.05 \\ 
     FO   & -0.07 & -0.07 & -0.09 & -0.06 & 1     & 0.1   & 0.13 \\ 
     \cellcolor{Gray}EoM  & \cellcolor{Gray}-0.45 & \cellcolor{Gray}-0.36 & \cellcolor{Gray}0.11  & \cellcolor{Gray}-0.47 & \cellcolor{Gray}0.62  & \cellcolor{Gray}1     & \cellcolor{Gray}0.82 \\ 
     TEoM & -0.5  & -0.38 & 0.13  & -0.45 & 0.86  & {\bf 0.86}  & 1 \\ \hline
    \end{tabular}
\end{minipage} \hspace{4cm}
\begin{minipage}{.4\textwidth}
	\caption{DS-based RCP.}
    \label{tab:ds_rcp}
    \begin{tabular}{|r|ccccccc|}
    \hline
           & PoB   & O     & PoR   & PoD   & FO    & EoM   & TEoM \\ \hline
     PoB   & 1     & {\bf -0.6}  & 0.55  & -0.14 & -0.54 & -0.48 & -0.57 \\ 
     \cellcolor{Gray}O     & \cellcolor{Gray}{\bf -0.02} & \cellcolor{Gray}1     & \cellcolor{Gray}-0.11 & \cellcolor{Gray}-0.16 & \cellcolor{Gray}-0.16 & \cellcolor{Gray}0.19  & \cellcolor{Gray}-0.15 \\ 
     PoR   & 0.65  & {\bf -0.33} & 1     & {\bf 0.45}  & -0.7  & -0.68 & -0.75 \\  
     \cellcolor{Gray}PoD   & \cellcolor{Gray}-0.06 & \cellcolor{Gray}-0.18 & \cellcolor{Gray}0.17  & \cellcolor{Gray}1     & \cellcolor{Gray}-0.18 & \cellcolor{Gray}-0.13 & \cellcolor{Gray}-0.19 \\ 
     FO    & -0.08 & -0.06 & -0.09 & -0.06 & 1     & 0.09  & 0.09 \\ 
     \cellcolor{Gray}EoM   & \cellcolor{Gray}-0.35 & \cellcolor{Gray}0.35  & \cellcolor{Gray}-0.42 & \cellcolor{Gray}-0.21 & \cellcolor{Gray}0.46  & \cellcolor{Gray}1     & \cellcolor{Gray}0.66 \\ 
     TEoM  & -0.16 & -0.1  & -0.17 & -0.12 & 0.34  & {\bf 0.24}  & 1 \\ \hline
    \end{tabular}
\end{minipage}
} 
\end{table}

The results of the macro RCP analysis for six of the relations we analyzed (Tab.\ref{tab:rcp}) shows that the $place\_of\_birth$ relation has a high causal power (0.64) over $origin$, meaning that when $place\_of\_birth$ is annotated in a sentence, $origin$ is also likely to appear, with the inverse causal power at 0.88.  This high co-causality seems to indicate a confusion between the two relations. Note also that these two relations have significant differences in causal power in the DS-based data. In contrast, $place\_of\_death$ has a high causal power over $places\_of\_residence$ in the DS data (0.45), reflecting the high error rate of $place\_of\_death$ caused by the overlap in the KB with $places\_of\_residence$.

In the crowd data we see a much higher co-causality for $employee\_or\_member$ and $top\_employee\_or\_member$, with only a slight preference in the data for what we expect to be the ``correct'' causal direction (that $top\_employee\_or\_member$ causes $employee\_or\_member$), but in the DS-based analysis, the incorrect interpretation drops a lot. In manual error analysis we observed that these are properties of the data set, which talk about more famous people who tend to be leaders and founders, not ``regular'' employees.  Table~\ref{tab:ds_fn_ex} in the Appendix shows several examples sentences with false negative DS labels due to missing causality.

All told, the differences (in bold in Tab.\ref{tab:rcp}) between the crowd and DS-based causal power accounts for some of the classification errors in our trained system, and we expect them to be a significant cause of error in systems that try to learn cross-relation signals from DS data alone.

Among the non-symmetric causal pairs we see that $top\_employee\_or\_member$ causes $founded\_organization$, $employee\_or\_member$ causes $founded\_org$, and $top\_employee\_or\_member$ causes $founded\_org$.  These again appear to be properties of the data set.

\section{Enhancing Distant Supervision with CrowdTruth}
We expect that the metrics from CrowdTruth annotation can be used to systematically enhance DS data at scale, without requiring the crowd to annotate the entire set.  As a preliminary headroom exercise, we ran three experiments to test a few simple heuristic characterizations of our analysis, and compared them to a baseline.  In each experiment, we changed only the DS training set (using the methods described below).  We used the data in our previously held-out test set as an evaluation target, again processing the continuous SRS scores with a threshold of 0.5 to yield discrete truth values for calculating P, R, and F.  Results are shown in Tab.~\ref{tab:cnn_res}.

\begin{enumerate}

\item {\bf DS:} The baseline to which the other experiments are compared. The per-relation training labels are binary based on the results of DS.

\item {\bf DS merged:} Based on the results of the causality analysis, the training set is augmented to reflect the highest cross-relation signals.  We merge relations with symmetric RCP ($origin$ and $place\_of\_birth$), and add the implied relation in the case of asymmetric RCP ($employee\_or\_member$  and $top\_employee\_or\_member$). To merge, the {\bf DS} baseline data is updated so that the symmetric relations always co-occur, and adding caused relation whenever the caused relation appears. This approach shows a huge improvement across the board over the baseline, with the overall highest P and F.

\item {\bf DS\_RCP:} Instead of manually identifying merged relations, the training data is augmented by using the RCP scores. When a relation $i$ has a positive {\bf DS} label for a given sentence, the labels of all other relations $j \neq i$ are updated by adding the macro RCP that $i$ has over $j$. The maximum value for the label is clipped at 1, to keep scores in the $[0,1]$ interval. The training labels in this set have continuous values, as opposed to the binary values in the previous two sets. The formula for updating the training label for relation $j$ in sentence $s$ is: $ DS\_RCP(s, i) = max[1, DS(s, j) + \sum_{i \neq j} RCP(i,j) \cdot DS(s,i) ],$ where $DS(s,i)$ is the DS label of relation $i$ in sentence $s$.  This method was comparable in precision to the baseline, but scored a huge win in recall.  The recall increase makes sense, though we have yet to investigate or explain the lack of increase in precision.

\item {\bf DS\_FP:} Our analysis showed that the $place\_of\_death$ relation was a large source of false positives in the DS data, because most of the positives were actually expressing $places\_of\_residence$.  In every sentence in the DS training set that had a 1 for $place\_of\_death$, we updated the score by subtracting its false positive ratio, which was used in the loss function as described above.  This did not impact the results over the baseline, mainly because there were not many $place\_of\_death$ relations in the DS data nor the test set, and any improvement did not impact the overall result.  We are confident that more systematic treatment of false positive rates will improve performance.

\end{enumerate}

\begin{table}[htb!]
\caption{Precision \& Recall at 20,000 training steps.}
\label{tab:cnn_res}
\centering
\begin{tabular}{r|ccc}
& {\bf Precision} & {\bf Recall} &{\bf F1 score} \\ \hline
{\bf DS} & 0.19 & 0.22 & 0.2 \\
{\bf DS merged} & {\bf 0.43} & 0.33 & {\bf 0.37}  \\
{\bf DS\_RCP} & 0.19 & {\bf 0.48} & 0.27 \\
{\bf DS\_FP} & 0.21 & 0.22 & 0.21  \\
\end{tabular}
\end{table}

\section{Discussion}

The preliminary results are not overwhelming, but highly indicative.  There is considerable headroom in cross-relation signals, and a more robust approach holds promise to eliminate manual analysis, and work as part of an overall pipeline that includes partial crowd data. We have shown a very significant variation in the false positive rate in distant supervision data, and it seems extremely likely that this can be exploited to improve training.

We are currently considering experiments that take advantage of another aspect of our CrowdTruth method: the identification of ambiguity in sentences where workers do not agree on the outcome.  We believe a more continuous truth measure as opposed to the rather arbitrary discrete measure will be productive. Finally, we are particularly excited about the possibility of using our approach in conjunction with logical reasoning approaches such as those reported in \cite{demeester2016regularizing}.  In this case, we are looking at more informed data that reflects human understanding and properties of the data set, to discover candidate relation pairs for investigating rules.

\section*{Appendix}

\begin{table}[bht!]
\centering
\caption{Example sentences with false positive $place\_of\_death$ and $origin$ DS labels due to multiple relations in the KB over $Person$ - $Location$ term types.}
\label{tab:ds_fp_ex}
\scalebox{0.85}{
\bgroup
\def\arraystretch{1.5}
\begin{tabular}{|p{8cm}|c|>{\centering\arraybackslash}p{1.7cm}|>{\centering\arraybackslash}p{1.3cm}|}
\hline
{\bf Sentence} & {\bf Relation} & {\bf Crowd SRS} & {\bf DS label} \\ \hline  \hline

\multirow{2}{8cm}{After growing up on Cat Island, {\bf Tony McKay} moved to {\bf New York City} at age 17 to study architecture.} & $place\_of\_death$ & 0.004 & 1 \\ \cline{2-4}
 & $places\_of\_residence$ & 0.995 & 1 \\ \hline

\multirow{2}{8cm}{The film is based very loosely on the lives of {\bf Wolfgang Amadeus Mozart} and Antonio Salieri, two composers who lived in {\bf Vienna, Austria}.} & $place\_of\_death$ & 0.074 & 1 \\ \cline{2-4}
 & $places\_of\_residence$ & 0.865 & 1 \\ \hline

\multirow{2}{8cm}{Marku Ribas is the side more Black music of this group and was {\bf Bob Marley}'s friend in the 1970s, {\bf Jamaica}, where he lived.} & $origin$ & 0 & 1 \\ \cline{2-4}
 & $places\_of\_residence$ & 0.87 & 1 \\ \hline

\multirow{2}{8cm}{{\bf Osama bin Laden} had moved from {\bf Saudi Arabia} to Sudan during the 1990-91 Gulf War.} & $origin$ & 0.3 & 1 \\ \cline{2-4}
 & $places\_of\_residence$ & 0.74 & 1 \\ \hline
 
\end{tabular}
\egroup
}
\end{table}

\begin{table}[thb!]
\centering
\caption{Example sentences with false negative $employee\_or\_member$ and $origin$ DS labels due to missing causal connections.}
\label{tab:ds_fn_ex}
\scalebox{0.85}{
\bgroup
\def\arraystretch{1.5}
\begin{tabular}{|p{7.5cm}|c|>{\centering\arraybackslash}p{1.7cm}|>{\centering\arraybackslash}p{1.3cm}|}
\hline
{\bf Sentence} & {\bf Relation} & {\bf Crowd SRS} & {\bf DS label} \\ \hline  \hline

\multirow{2}{7.5cm}{China on Monday officially appointed {\bf Donald Tsang} as {\bf Hong Kong}'s chief executive for a second term.} & $employee\_or\_member$ & 0.623 & 0 \\ \cline{2-4}
 & $top\_employee\_or\_member$ & 0.753 & 1 \\ \hline

\multirow{2}{7.5cm}{More than 3,000 taxi drivers blocked {\bf Rome}'s historic centre Wednesday to protest extra licences given by mayor {\bf Walter Veltroni}.} & $employee\_or\_member$ & 0.529 & 0 \\ \cline{2-4}
 & $top\_employee\_or\_member$ & 0.841 & 1 \\ \hline

\multirow{2}{7.5cm}{Early years {\bf Joey Harrington} was born and raised in {\bf Portland, Oregon}, where he has resided his entire life.} & $origin$ & 0.645 & 0 \\ \cline{2-4}
 & $place\_of\_birth$ & 0.867 & 1 \\ \hline

\multirow{2}{7.5cm}{{\bf Nelli Zhiganshina} (born March 31, 1987 in {\bf Moscow}, Russia) is a Russian ice dancer who currently represents Germany.} & $origin$ & 0.555 & 0 \\ \cline{2-4}
 & $place\_of\_birth$ & 0.791 & 1 \\ \hline
 
\end{tabular}
\egroup
}
\end{table}

\bibliographystyle{abbrv}
\bibliography{sources}
\end{document}